\documentclass[letterpaper]{article} 
\usepackage{aaai24}  
\usepackage{times}  
\usepackage{helvet}  
\usepackage{courier}  
\usepackage[hyphens]{url}  
\usepackage{graphicx} 
\usepackage{natbib}  
\usepackage{caption} 
\usepackage{algorithm}
\usepackage{algorithmic}
\usepackage{amsmath}
\usepackage{subfigure}
\usepackage{booktabs}  
\usepackage{threeparttable} 
\usepackage{multirow}
\usepackage{bbding}
\usepackage{color}
\usepackage{amssymb}
\usepackage{bm}
\usepackage{makecell}
\usepackage{adjustbox}
\usepackage{newfloat}
\usepackage{listings}
\DeclareCaptionStyle{ruled}{labelfont=normalfont,labelsep=colon,strut=off} 
\floatstyle{ruled}
\newfloat{listing}{tb}{lst}{}
\floatname{listing}{Listing}

\title{DOCTR: Disentangled Object-Centric Transformer 
for Point Scene Understanding}
\author{
    Xiaoxuan Yu\textsuperscript{\rm 1}\thanks{Corresponding author},
    Hao Wang\textsuperscript{\rm 1},
    Weiming Li\textsuperscript{\rm 1},
    Qiang Wang\textsuperscript{\rm 1},
    Soonyong Cho\textsuperscript{\rm 2},
    Younghun Sung\textsuperscript{\rm 2}
}
\affiliations{

    \textsuperscript{\rm 1}Samsung Research China – Beijing \\
    \textsuperscript{\rm 2}Samsung Advanced Institute of Technology\\
    xiaoxuan1.yu@samsung.com, hao1.wang@samsung.com, weiming.li@samsung.com, qiang.w@samsung.com,
    soonyong.cho@samsung.com, younghun.sung@samsung.com
}

\usepackage{bibentry}

\begin{document}

\maketitle

\begin{abstract}
Point scene understanding is a challenging task to process real-world scene point cloud, which aims at segmenting each object, estimating its pose, and reconstructing its mesh simultaneously. Recent state-of-the-art method first segments each object and then processes them independently with multiple stages for the different sub-tasks. This leads to a complex pipeline to optimize and makes it hard to leverage the relationship constraints between multiple objects. In this work, we propose a novel Disentangled Object-Centric TRansformer (DOCTR) that explores object-centric representation to facilitate learning with multiple objects for the multiple sub-tasks in a unified manner. Each object is represented as a query, and a Transformer decoder is adapted to iteratively optimize all the queries involving their relationship. In particular, we introduce a semantic-geometry disentangled query (SGDQ) design that enables the query features to attend separately to semantic information and geometric information relevant to the corresponding sub-tasks. A hybrid bipartite matching module is employed to well use the supervisions from all the sub-tasks during training. Qualitative and quantitative experimental results demonstrate that our method achieves state-of-the-art performance on the challenging ScanNet dataset. Code is available at https://github.com/SAITPublic/DOCTR.
\end{abstract}
\section{Introduction}


Understanding 3D scene is important for various spatial modeling and interaction applications such as augmented reality (AR), autonomous driving, and robotics. With the progress of 3D deep learning, recent real-world applications seek more comprehensive understanding of rich and detailed attributes for all objects of interests in a scene. This paper addresses a recent task of point scene understanding \cite{nie2021rfd, tang2022point} that simultaneously involves several sub-tasks including object classification, instance segmentation, pose estimation, and object mesh reconstruction. The input is point cloud of real-world scene that is obtained by 3D scanning or reconstruction. Such data often contains noisy and missing portions due to occlusions or sensor limitations, which make the task even more challenging.
      

  \begin{figure}[t]
   \begin{center}
      \includegraphics[width=0.9\linewidth]{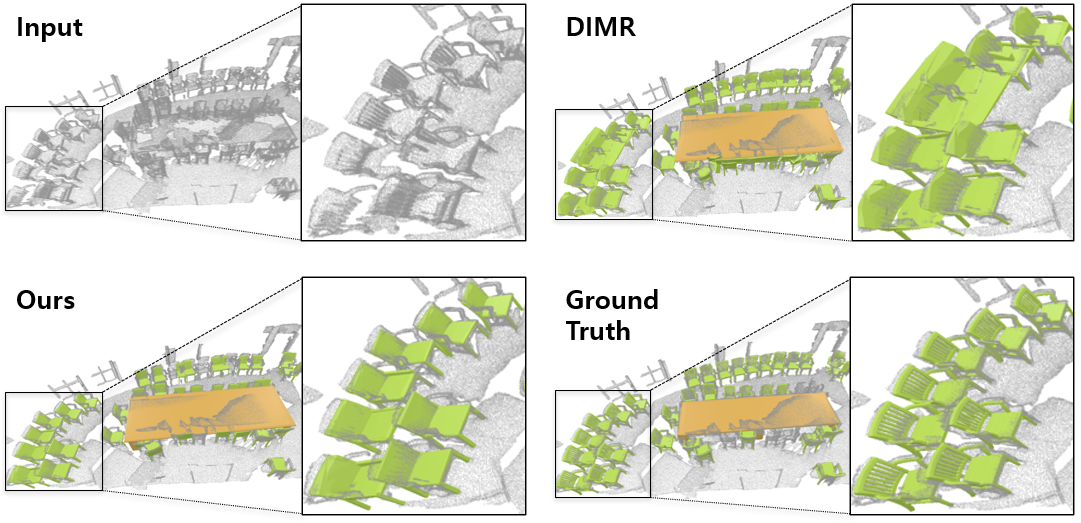}
   \end{center}
       \caption{Point Scene Understanding. With an incomplete point cloud scene as input, our method learns to segment each object instance and reconstruct its complete mesh. Compared to the recent DIMR \cite{tang2022point}, our proposed DOCTR has a compact pipeline and achieves more accurate results in challenging cases such as when multiple objects are in close proximity.}
   \label{comparasion}
   \end{figure}
 
For the point scene understanding task, an early method RfD-Net\cite{nie2021rfd} is among the first to learn object meshes at semantic-instance level directly from points. RfD-Net proposes a reconstruction-from-detection framework that enables identification and reconstruction of object meshes at a high resolution. It confirms that object recognition and reconstruction are mutually reinforcing tasks. Recently, DIMR \cite{tang2022point} founds that the reconstruction-from-detection pipeline tends to fail when reconstructing high fidelity objects. DIMR shows that replacing the detection backbone in RfD-Net by a segmentation backbone and disentangling instance completion and mesh generation contribute to improve object reconstruction quality. These designs empower DIMR to achieve state-of-the-art (SOTA) performance. However, DIMR still relies on geometric clustering for object segmentation that requires tuning of hyper-parameters such as the radius for clustering. Meanwhile, after object segmentation, DIMR 
treats each object instance independently to estimate its pose, shape latent code, and reconstruct its mesh. DIMR's pipeline is complex to jointly optimize the segmentation and other sub-tasks. Further, the separate processing for each object makes it hard to leverage the relationship between multiple objects. In test with real-world scenes, DIMR is prone to artifacts for multiple nearby objects as shown in Figure \ref{comparasion}.



To solve the above issues, our work is inspired by object-centric learning \cite{carion2020end,locatello2020object}, which proposes to learn efficient representations of a scene by decomposing it into objects, thus is natural to explore the object relationship. Recently, object-centric learning methods are introduced into 3D point cloud segmentation task \cite{Schult23ICRA, sun2023superpoint} and show promising performance. In this paper, we attempt to introduce an object-centric learning based method for the point scene understanding task. Specifically, each object is represented as a query, and a Transformer decoder is adapted to iteratively
optimize all the queries involving their relationship.




However, applying an object-centric framework to the point scene understanding task is not straight forward. In existing methods such as RfD-Net and DIMR, after obtaining object proposals, each object's point cloud is transformed to object's canonical coordinate frame using the initial pose estimation result. This 3D alignment is important to improve subsequent geometric sub-tasks such as object pose refinement, shape completion, and mesh reconstruction. However, object-centric Transformer uses a unified representation for each object, which is not meaningful to perform any 3D coordinate frame alignment. This makes it difficult to learn object geometric information such as pose and shape. To address this issue, we propose a semantic-geometry disentangled query (SGDQ) design. Different from the origin query, SGDQ disentangles each query to a semantic part and a geometry part. The semantic part is supervised by semantic sub-tasks such as object classification and semantic instance segmentation. The geometric part is supervised by the geometric sub-tasks such as pose estimation and mesh reconstruction. With the distinct focuses on semantic and geometric sub-tasks, our SGDQ empowers disentangled learning for the task-specific features. 

Following the above, we propose our method as Disentangled Object-Centric TRansformer (DOCTR). Our DOCTR consists of a backbone, a disentangled Transformer decoder with the SGDQ design, a prediction head for the sub-tasks of point scene understanding, and a shape decoder. It facilitates learning with multiple objects for the multiple sub-tasks in a unified manner. Compared to DIMR, our DOCTR has a much more compact pipeline as shown in Figure 2. Meanwhile, different from  Mask3D \cite{Schult23ICRA}, our SGDQ design allows learning for multiple different sub-tasks that relates to either semantic or geometric attributes of the scene objects. For training our DOCTR model, 
we employ a hybrid bipartite matching strategy during the matching process between ground truths and SGDQs. We also propose a mask-enhanced box refinement module that leverages segmentation to improve pose estimation. Extensive experiments are performed with the real-world large-scale ScanNet dataset in comparison to the SOTA methods.

 The contributions of this paper are as follows:
  \begin{itemize}
   \item As far as we know, our DOCTR is the first to introduce an object-centric Transformer-based network for the point scene understanding task that allows learning with multiple objects and multiple sub-tasks in a unified manner.
   \item We propose semantic-geometry disentangled query (SGDQ) that enables the DOCTR network to extract semantic and geometric features for the different sub-tasks to effectively use disentangled representations. 
   
   \item Qualitative and quantitative experimental results on the challenging ScanNet dataset show that our proposed method achieves superior performance than previous SOTA methods, especially for the challenging cases such as cluttered scenes with nearby objects.
   
  \end{itemize}
  \begin{figure}[t]
   \begin{center}
      \includegraphics[width=1.0\linewidth]{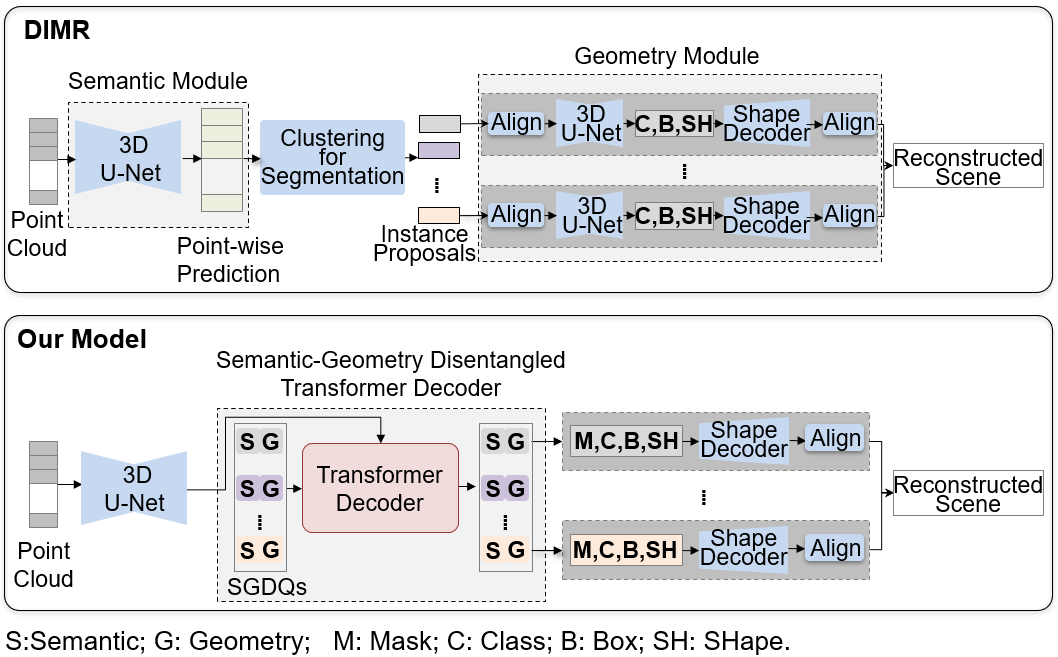}
   \end{center}
      \caption{Comparison of the pipelines between DIMR \cite{tang2022point} and our DOCTR. DIMR first segments each object instance by clustering and then processes them independently with multiple stages for the different sub-tasks. Our DOCTR pipeline is much more compact that represents each object instance as a query and optimize for multiple objects and multiple sub-tasks in a unified manner.}
   \label{fig1}
   \end{figure}
 
\begin{figure*}[htbp]
   \begin{center}
      \includegraphics[width=0.8\linewidth]{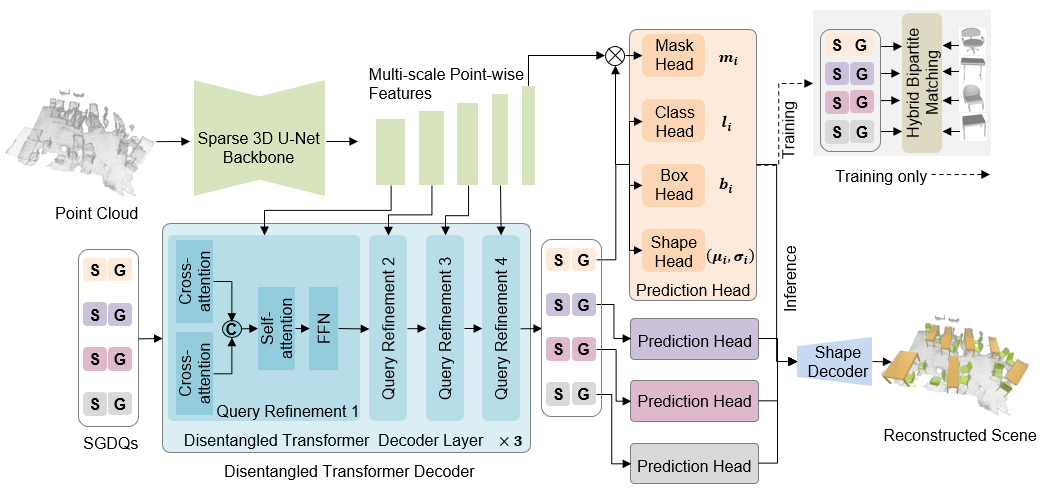}
   \end{center}
      \caption{Our proposed DOCTR pipeline consists of a backbone, a disentangled Transformer decoder (DTD), a prediction head, and a shape decoder. We propose semantic-geometry disentangled queries (SGDQs) to represent scene objects. The DTD is trained to attend the SGDQs to multi-scale point-wise features extracted from the sparse 3D U-Net backbone. In inference, each object SGDQ is passed to the prediction head to predict the object's mask, class, box (pose), and shape code. The shape code is input to the shape decoder to reconstruct the object's complete mesh, which is then aligned by the estimated pose to the scene. }
   \label{fig1}
   \end{figure*}
  \section{Related Work}
\noindent{\textbf{Point scene understanding.} }
Many real-world applications require understanding both semantic and geometric attributes of 3D scenes. Among different types of inputs, point scene understanding is related to 3D deep learning using only scene's point cloud as input. While point cloud is commonly used for object-level tasks such as object completion \cite{huang2019deep,wang2020cascaded,xie2020grnet,yuan2018pcn},
less work \cite{zhong2020semantic,rist2020scssnet,rist2021semantic} use point cloud with neural networks for scene understanding. 
Early works \cite{garbade2019two,roldao2020lmscnet,wu2020scfusion,yan2021sparse} often use voxelized input and then predict the semantic label of each voxel in both visible and occluded regions. They aim to jointly estimate the complete geometry and semantic labels from partial input. Working with voxels, computational cost and resolution need to be balanced which limits to reconstruct high fidelity object meshes. 




To solve such issues, recent point scene understanding methods explore the relationship of semantic and geometric information, containing object localization and reconstruction. RfD-Net \cite{nie2021rfd} is among the first to propose a reconstruction-from-detection framework to detect objects 
and generate complete object meshes. It shows that the tasks of object
detection and object completion are complementary. Current state-of-the-art is achieved by DIMR \cite{tang2022point} that proposes to use segmentation network
instead of detection network to improve the accuracy of object recognition. DIMR also disentangles the tasks of
object completion and mesh generation, mitigating
the ambiguity of learning complete shapes from incomplete point cloud observations. In contrast to existing methods, our proposed approach introduces an object-centric Transformer-based network for the point scene understanding task that enables simultaneous learning with multiple objects and multiple sub-tasks in a unified manner.

\vspace{0.5cm}

\noindent{\textbf{Object-centric learning.}} 
Learning with object-centric representation is a promising approach to understanding real-world complex scenes with multiple objects. 
With the help of set prediction formulation, DETR\cite{carion2020end} proposes to learn object-centric representations
with the Transformer architecture for 2D object detection. In \cite{locatello2020object}, a fully-unsupervised approach based on slot attention for object-centric learning is proposed. Object-centric learning has achieved great success in 2D tasks\cite{zhu2020deformable, liu2022dab, zhang2022dino}.
Recently, there have been notable advancements in applying object-centric learning to 3D point cloud segmentation or detection tasks\cite{Schult23ICRA, sun2023superpoint, zhu2023conquer}. These methods have demonstrated performance that rivals or even surpasses the current SOTA approaches in these fields.
Different from the recent object-centric models for 3D tasks such as Mask3D \cite{Schult23ICRA}, our work proposes a semantic-geometry disentangled query design that allows object-centric Transformer to deal with multiple different semantic and geometric sub-tasks simultaneously.



\section{Methods}


An overview of our posed DOCTR pipeline is illustrated in Figure \ref{fig1}. Our DOCTR consists of a sparse 3D U-Net backbone,  a disentangled Transformer decoder (DTD), a prediction head, and a shape decoder. All objects are represented by our designed SGDQs and each SGDQ corresponds to an object instance. During training, the Transformer decoder is trained to attend the SGDQs to the multi-scale point-wise features extracted by the backbone. To supervise the training, our proposed hybrid bipartite matching is used to assign either a `no object' class or an object's ground-truth attributes to the corresponding object SGDQ. In inference, for each object's SGDQ, the prediction head predicts its mask, class, pose (box), and a shape code. The predicted shape code is decoded by a shape decoder to reconstruct the object's complete mesh. Then the mesh is transformed by the estimated pose to align to the scene's coordinate frame as the object's reconstruction. All the reconstructed objects comprise the final reconstructed scene. Next we'll describe the details for each component of the pipeline.

\subsection{Sparse 3D Backbone}

   From the input point cloud ${\bf P}\in \mathbb{R}^{N\times3}$, we employ a sparse 3D U-Net \cite{choy20194d} to extract multi-scale point-wise features ${\{\bf F}_l\}\in\mathbb{R}^{N_l\times D^f_l}$, where $l=0,\dots,L$ is scale level, $N_l$ and $D^f_l$ are spatial dimension and  feature dimension of ${\bf F}_l$ respectively. Especially, the spatial dimension of ${\bf F}_0$ and $\bm{P}$ are the same. Different from DIMR, our method does not add any additional semantic branch or offset branch for instance segmentation.
   

   \subsection{Disentangled Transformer Decoder}
   

   Point scene understanding comprises sub-tasks involving different types of information. Specifically, the sub-tasks of instance segmentation and classification are more related to semantic information.  Meanwhile, the sub-tasks of pose estimation and shape reconstruction are more related to geometry information. To ensure optimal learning for all the sub-tasks, we propose disentangled Transformer decoder (DTD), which aims to decouple the learning of semantic features and geometric features. This allows our model to learn the most relevant information for each sub-task. 
   
   \noindent{\textbf{Semantic-geometry disentangled query.}} As a key design of the DTD, we pose semantic-geometry disentangled query (SGDQ). All scene objects are represented by a set of SGDQs and each SGDQ corresponds to one object instance. Unlike the query used in the vanilla object-centric framework, the feature of each SGDQ is divided into a semantic part and a geometric part. In training, the disentanglement of semantic and geometric parts is achieved through separate supervision with the different sub-task prediction heads. The semantic part is supervised by the sub-tasks of object classification and semantic instance segmentation. The geometric part is supervised by the sub-tasks of pose estimation and mesh reconstruction. 
  We choose the same fixed number of SGDQs for all scenes, whose value is often much larger than the number of objects in the scene. SGDQs are randomly initialized and iteratively learned to be representations of objects in the scene. 
\begin{figure}[t]
   \begin{center}
      \includegraphics[width=0.8\linewidth]{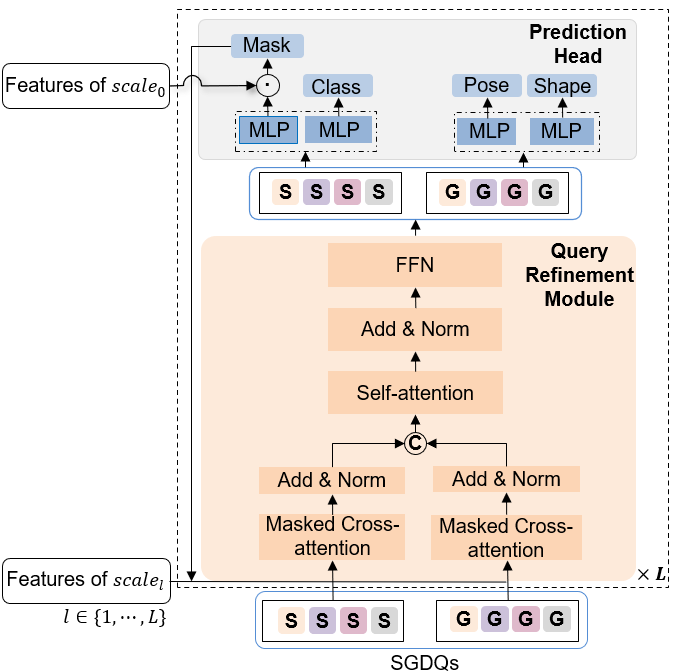}
   \end{center}
      \caption{Disentangled Transformer Decoder Layer}
   \label{fig2}
   \end{figure}
   
   \noindent{\textbf{Disentangled transformer decoder layer.}}
    Our DTD  consists of multiple DTD layers, and each DTD layer contains multiple query refinement modules. Each query refinement module corresponds to a different level of the multi-scale features ${\{\bf F}_l\}$ for $l=1,\dots,L$ as shown in Figure \ref{fig2}.
    The semantic parts and geometric parts of SGDQs are separately refined by cross-attending to the scene's point-wise features and fused at the object level through self-attention. Starting from $M$ randomly initialized SGDQs, through the series of query refinement modules, the SGDQs are optimized in a coarse-to-fine manner progressively. In our model, we use three DTD layers and four levels ($\bm{L}=4$) in each DTD layer. 
    
    Different from the original Transformer decoder layer, our DTD layer is designed to learn disentangled semantic and geometric representations of SGDQs. At each level $l$ in a decoder layer, each SGDQ competes for attending to the features ${\bf F}_l$ from one object instance through attention mechanism. We denote ${\bf Q}^{l}$ as an input SGDQ of level $l$, and  ${\bf Q}^{l+1}$ as its output SGDQ. Each SGDQ is composed of a semantic part and a geometric part, denoted as $ {\bf Q}^{l} = ({\bf Q}^{l}_s,{\bf Q}^{l}_g)$.
   In each query refinement module of DTD layer, we firstly utilize two cross-attention modules to learn the semantic and geometric information separately from the multi-scale point-wise features at level $l$. In the semantic aware cross-attention module, we first apply learnable linear transformations, denoted as ${\bf q}_s, {\bf k}_s, {\bf v}_s$ respectively, to features ${\bf F}_p$ and semantic part queries ${\bf Q}_s^l$ to obtaining query ${\Bar{\bf Q}}_s^l\in\mathbb{R}^{M\times D}$, key ${\bf K}_s^l\in\mathbb{R}^{N\times D}$ and value ${\bf V}_s^l \in\mathbb{R}^{N\times D}$. Then matrix multiplication between ${\hat{\bf Q}}_s^l$ and ${\bf K}_s^l$ gives the correlation matrix between $M$ queries and $N_l$ input features. Here we use the masked cross-attention where each query only attends to features within binary instance mask ${\bf m}$ predicted by the mask prediction head using ${\bf Q}_s^{l-1}$, which can be formulated as:
   \begin{equation}
   \hat{{\bf Q}}_s^{l+1} = \text{softmax}(\overbrace{{\bf q}_s({\bf Q}^{l}_s)}^{\Bar{\bf Q}_s^l}\cdot\overbrace{{\bf k}_s({\bf \Tilde{m}}({\bf F}, {\bf m}))}^{{\bf K}_s^l} {}^T)\overbrace{{\bf v}_s({\bf F})}^{{\bf V}_s^l},
   \end{equation}
   where 
    \begin{equation}
        {{\bf \Tilde{m}}({\bf F}, {\bf m})}_{j,k} = \left\{
        \begin{aligned}
            &{\bf F}_{j,k}, &\quad  {\bf m}_{j,k} = 0,\\
            &0, &\quad {\bf m} _{j,k}=1.
        \end{aligned}
        \right.
    \end{equation}
   The same attention method is used in the geometric part ${\bf Q}_g^l$:
   \begin{equation}
   \hat{{\bf Q}}_g^{l+1} = \text{softmax}({\bf q}_g({\bf Q}^{l}_g)\cdot{\bf k}_g({\bf \Tilde{m}}({\bf F}, {\bf m}))^T){\bf v}_g({\bf F}).
   \end{equation}
   These two cross-attentions empower the queries to extract semantic and geometric information from the point-wise features. Then, the semantic and geometric features will be concatenated and fed into a self-attention module and a Feed Forward Network (FFN) to gather  context information:
    \begin{equation}
   \begin{aligned}
      \label{eq1}
      &\hat{{\bf Q}}^{l+1}= \text{self-attention}((\hat{{\bf Q}}_s^{l+1},\hat{{\bf Q}}_g^{l+1})),\\
      &{\bf Q}^{l+1} = \text{FFN}(\hat{{\bf Q}}^{l+1}).
   \end{aligned}
   \end{equation}

\noindent{\textbf{Prediction head.}} After DTD, each SGDQ $[{\bf Q}]_i$ for $i=0,\cdots,M-1$ attends to the features of a specific object instance. These SGDQs are then fed into a prediction head whose weights are shared across queries. The prediction head consists of four MLP-based task heads, including mask head, class head, box head, and shape head. The mask head takes semantic part $[{\bf Q}_s]_i$ and the point-wise features from the sparse 3D U-Net as input. It maps $[{\bf Q}_s]_i$ through an MLP $f_{mask}(\cdot)$, and outputs object binary mask ${\bf m}_i$ by computing dot products between the mapped query features and point-wise features ${\bf F}_0$ with a threshold at 0.5:
\begin{equation}
    [{\bf m}]_i = \{\sigma ({\bf F}_0 \cdot f_{\text{mask}}([{\bf Q}_s]_i){}^T)>0.5 \in \{0,1\}^{N}\}.
\end{equation}
The class head also takes the semantic part $[{\bf Q}_s]_i$ as input, and outputs classification logits ${\bf l}_i\in\mathbb{R}^{C}$  for each instance, where $C$ is the total class number. 

The box head takes the geometric part $[{\bf Q}_g]_i$ as input, and outputs box vector $\bm{b}_i$, including rotation angle $r_i\in[-\pi,\pi]$ along the z-axis following \cite{nie2021rfd}, center of bounding box ${\bf c}_i=(x_i,y_i,z_i)$ and size of bounding box ${\bf s}_i=(s^x_i,s^y_i,s^z_i)$ and IoU score $\text{score}^{\text{IoU}}_i$, denoted as $\bm{b}_i = (r_i, \bm{c}_i, \bm{s}_i, \text{score}^{\text{IoU}}_i)$.

For shape completion, we assume that the complete shape is sampled from a latent Gaussian distribution and learn it through the reparameterization trick \cite{kingma2013auto}. 
The shape head takes the geometric part $[{\bf Q}_g]_i$ as input and regresses mean and standard deviation ${\bm \mu}_i,{\bm \sigma}_i\in\mathbb{R}^{D_{\text{shape}}}$ for the shape latent Gaussian distribution, where $D_{\text{shape}}$ is the latent shape code dimension. We decode the latent code $\bm{z}_i \sim \mathcal{N}(\bm{\mu}_i,\bm{\sigma}_i)$ to mesh  only at inference time through a pretrained shape decoder of BSP-Net\cite{chen2021learning, chen2020bsp}. Details of the shapde decoder can be found in \cite{tang2022point}.

\subsection{Training Design} 
\noindent{\textbf{Hybrid bipartite matching.}} During training, we utilize ground truth annotations for each object instance, including the mask, class, box, and mesh. For determining the correspondence between the queries and the ground truths, we use bipartite matching following DETR \cite{carion2020end}. For each SGDQ, a pair of mask and box predicted by two task heads may be inconsistent with each other. To obtain accurate and consistent matching, we employ a mixed cost of predicted mask, box, and class when performing bipartite matching. We construct a cost matrix $\mathcal{C}$ defined as follows:
 \begin{equation}
\mathcal{C} = \lambda_1\mathcal{L}^\text{mask}_\text{dice} + \lambda_2\mathcal{L}^\text{box}_\text{GIoU} + \lambda_3\mathcal{L}^\text{class} = \{\mathcal{C}_{jl}\},
\label{cost}
 \end{equation}
 in which $\mathcal{C}_{jl}$ is the similarity of the j-th query and the l-th ground truth. We set the weights to $\lambda_1 = 5$ and $ \lambda_2 =\lambda_3= 2$.
The meaning of each component will be explained in the upcoming discussion of training losses. After matching, those queries that are not assigned to any ground truth are given to the `no object' category and only participate in the classification task. Benefiting from the one-to-one matching mechanism, each object instance in the scene will be represented by only one query, which allows our network to directly perform sparse prediction.

\noindent{\textbf{Training losses.}} 
The loss $\mathcal{L}$ 
consists of semantic loss $\mathcal{L}^\text{sem}$ and geometric loss $\mathcal{L}^\text{geo}$.

Semantic tasks are supervised by :
\begin{equation}
\mathcal{L}^\text{sem} = \mathcal{L}^\text{mask}_\text{BCE} + \mathcal{L}^\text{mask}_\text{dice} + \mathcal{L}^\text{class},
 \end{equation}
 in which $\mathcal{L}^\text{mask}_\text{BCE}$ is the binary cross-entropy loss for mask, $\mathcal{L}^\text{mask}_\text{dice}$ is the Dice loss for mask,  and $\mathcal{L}^\text{class}$ is the cross-entropy loss for object-wise semantic label.
 
Geometry tasks are supervised by :
\begin{equation}
\mathcal{L}^\text{geo} = \mathcal{L}^\text{box}_\text{center} + \mathcal{L}^\text{box}_\text{size} + \mathcal{L}^\text{box}_\text{angle} + \mathcal{L}^\text{box}_\text{GIoU} + \mathcal{L}^\text{shape},
 \end{equation}
 where $\mathcal{L}^\text{box}_\text{center}$ is the Huber loss for bounding box center,  $\mathcal{L}^\text{box}_\text{size}$ is the Huber loss for bounding box size, $\mathcal{L}^\text{box}_\text{angle}$ contains 
 the cross-entropy loss for angle label and Huber loss for residual angle, $\mathcal{L}^\text{box}_\text{GIoU}$ is the 3D GIoU loss for box regression, $\mathcal{L}^\text{shape}$
 is the latent shape distribution loss for predicted shape latent distribution. The ground-truth latent shape distributions are obtained from a shape encoder of BSP-Net pretrained with ground-truth meshes. 
\begin{table*}[htbp]

	\centering
	\fontsize{9pt}{10pt}\selectfont
 \begin{adjustbox}{max width=\textwidth}
		\begin{tabular}{c|cccccc|c}
			\toprule
			&IoU@0.25
			&IoU@0.5&CD@0.1&CD@0.047&LFD@5000&LFD@2500&PCR@0.5\cr
			\midrule
			{RfD-Net\cite{nie2021rfd}}
			& 42.52 &14.35&46.37&19.09&28.59&7.80&43.49\cr
   {RfD-Net(refined data)}
			& 45.84 &15.60&46.97&18.41&29.90&6.83&43.49\cr
			{DIMR\cite{tang2022point}}
			& 46.34 &12.54&52.39&25.71&29.47&8.55&56.76\cr
               {DIMR(refined data)}
			& 53.18 &13.61&54.20&26.99&33.14&8.47&56.76\cr
                \hline
               {DOCTR w/o MEBR (ours)}
			& 55.18 &17.88&56.82&29.17&33.17&\bf{12.00}&55.70\cr
                {DOCTR w/ MEBR (ours)}
			& \bf{58.25} &\bf{19.60}&\bf{59.61}&\bf{31.46}&\bf{33.61}&10.9&\bf{59.89}\cr
			\bottomrule
		\end{tabular}
  \end{adjustbox}
 \caption{Comparisons on mesh completion quality and mapping quality. We report mean average precision
for different metric@threshold. For IoU and PCR, higher thresholds are more difficult. For
CD and LFD, smaller thresholds are more difficult. \label{mesh}}
\end{table*}
 \noindent{
 \subsection{Mask Enhanced Box Refinement}
 For an object instance $i$, it is observed that when the network predicts its mask $m_i$ accurately, the box $\bm{B}^m_i=(\bm{c}_i^p,\bm{s}_i^p)$ calculated from the object's point cloud tends to be more accurate than the network's predicted box $\bm{B}^p_i = (\bm{c}_i^m,\bm{s}_i^m)$. We consider the mask $m_i$ is ``sufficiently accurate",  when the distance $d(\bm{B}^p_i, \bm{B}^m_i)=|\bm{s}^{p}_i-\bm{s}^{m}_i|_{\infty}$ between $\bm{B}^m_i$ and $\bm{B}^p_i$ is less than $d_0 = 0.1\text{m}$.
 For each object point cloud $\bm{o}_i$, we apply the estimated rotation angle $r_i$ to transform it to a canonical coordinate system for mask enhanced box refinement (MEBR). For MEBR, denote the box calculated from the object point cloud $\bm{o}_{c}$ in canonical coordinate system as $\bm{B}^m_i$ and the box predicted by network as $\bm{B}^p_i$, then the refined box $\bm{B}^f_i$ is as follows:
 \begin{equation}
             \bm{B}^f_i = \left\{
        \begin{aligned}
            &\bm{B}^p_i, &\quad  \text{if} \quad d(\bm{B}^p_i, \bm{B}^m_i) \leq d_0,\\
            &\bm{B}^m_i, &\quad\text{otherwise}.
        \end{aligned}
        \right.
 \end{equation}

 \vspace{0.5cm}

 \section{Experiment}

 \subsection{Experimental Settings}
 \begin{table}[htbp]
	\centering
	\fontsize{9}{10}\selectfont
	\begin{threeparttable}
		
		\begin{tabular}{c|cc}
			\toprule
			&Prec.@0.25
			&Prec.@0.5\cr
			\midrule
			{RfD-Net(refined data)}
			& 22.99 &7.92\cr
			{DIMR(refined data)}
			& 36.23 &11.91\cr
   \hline
			{DOCTR w/o MEBR (ours)}
			& 47.75 &21.08\cr
			{DOCTR w/ MEBR (ours)}
			& \bf{49.14} &\bf{21.63}\cr
			\bottomrule
		\end{tabular}
	\end{threeparttable}
 \caption{Object Recognition Precision.\label{prec} We report the precision of object
recognition at different IoU thresholds. }
\end{table}
 \textbf{Datasets.} As previous works for point scene understanding, our experiments are conducted with the ScanNet V2 dataset \cite{dai2017scannet} (abbreviated as ScanNet for short). 
 The ScanNet dataset is a richly-annotated dataset that contains point clouds of 1513 real world indoor scenes labeled at instance level. 
The Scan2CAD dataset \cite{avetisyan2019scan2cad} provides additional geometric instance annotations by aligning CAD models from ShapeNet \cite{chang2015shapenet} to instances in ScanNet.
To deal with data inconsistency issue, DIMR \cite{tang2022point} relabels the datasets based on a compatible label system including 8 object categories. We follow exactly the same data split and pre-processing method as DIMR. A minor exception is when searching for the best matches between CAD models in Scan2CAD and instances in ScanNet based on box IoU, we further require that the matched pairs belong to the same category. This excludes some cases of mismatching caused by overlapping object boxes. The annotation variations will be made publicly available. For fair comparison, we re-evaluate previous RfD-Net and DIMR models on the refined annotations. 
%

\noindent{\textbf{Metrics.}}
Following previous works \cite{nie2021rfd,tang2022point}, we evaluate performance from two aspects including completion quality and mapping quality.

\noindent(i) \emph{Completion quality}. It is measured by the distance between the reconstructed meshes and the aligned models in the ground truth. We employ three metrics to evaluate completion quality following \cite{tang2022point}: the voxel-based 3D Intersection over Union (IoU) metric, the point-based Chamfer Distance (CD) metric, and the mesh-based Light Field Distance (LFD) metric. 
We adopt these metrics with the same thresholds as these in DIMR to determine whether a predicted mesh can match a ground-truth mesh. We also report the mean precision of IoU over all classes.


\noindent(ii) \emph{Mapping quality}. 
We adopt the Point Coverage Ratio (PCR) proposed in \cite{tang2022point} to measure the distance between the reconstructed mesh and the original point cloud, which computes the nearest distance from each observed instance point to the corresponding mesh surface.

\noindent{\textbf{Implementation details.}} The Minkowski Res16UNet34C \cite{choy20194d}  is used as the 3D U-Net backbone. The SGDQ in our DTD is represented by 256D features, of which the first 128 dimensions represent semantic features, and the latter 128 dimensions represent geometric features.  Following  \cite{sun2023superpoint}, we also reduce memory consumption by computing the dot product between instance queries and aggregated point features within segments which are obtained from a graph-based segmentation \cite{felzenszwalb2004efficient}.  During training, we use the AdamW \cite{loshchilov2017decoupled} optimizer for 600 epochs with a batch size of 5 on a single Nvidia RTX A6000 GPU for all the experiments.  One-cycle learning rate schedule \cite{smith2019super} is utilized with a maximum learning rate of $10^{-4}$ and a minimum learning rate of $10^{-6}$.  Standard data augmentation are performed on point cloud including horizontal flipping, random rotations around the z-axis, elastic distortion, and random scaling. 

\subsection{Comparisons to the State-of-the-Arts}
Table \ref{mesh} shows the quantitative comparisons between our method with RfD-Net \cite{nie2021rfd} and DIMR \cite{tang2022point} in metrics of completion quality and mapping quality. As described in previous section, some wrong annotations are removed, so we use the officially released models of RfD-Net and DIMR to re-evaluate the metrics for fair comparison. For reader's convenience, Table \ref{mesh} includes both the metrics from the original paper report and our re-evaluation results. We provide results for our DOCTR method in two versions, one is with  mask enhanced box refinement (MEBR) and the other is without MEBR. As shown in Table \ref{mesh}, our DOCTR (w/o MEBR) achieves improved performance on most metrics. Especially in the IoU@0.5, our DOCTR (w/o MEBR) surpasses RfD-Net by a large margin, while DIMR shows slightly worse performance than RfD-Net. Also, our DOCTR (w/o MEBR) obtains consistent improvements on all the CD and LFD metrics.   
A reason for the lower PCR@0.5 is that the PCR metric is highly related to point cloud and a slight box deviation can lead to a significant decrease, even if our reconstructed shape is of good quality. With adding the MEBR to enhance box quality, our DOCTR (with MEBR) achieves further improvement and surpasses DIMR in all the metrics including the PCR.  
\begin{figure*}[t]
    \centering
    \subfigure[DIMR]{
        \includegraphics[width=0.24\textwidth]{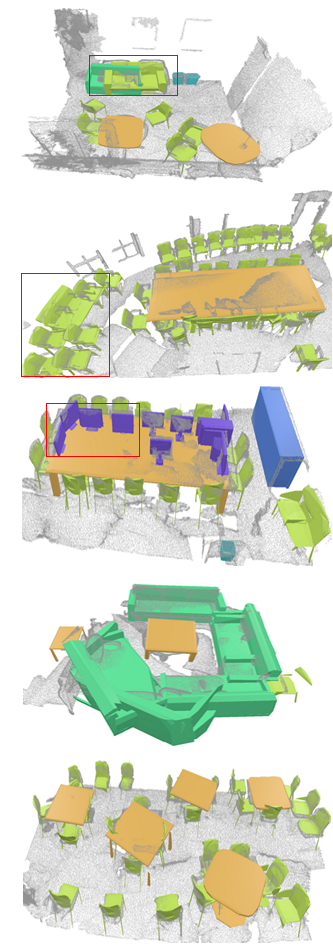}
        \label{DIMR}
    }
    \quad
    \subfigure[Ours]{
        \includegraphics[width=0.24\textwidth]{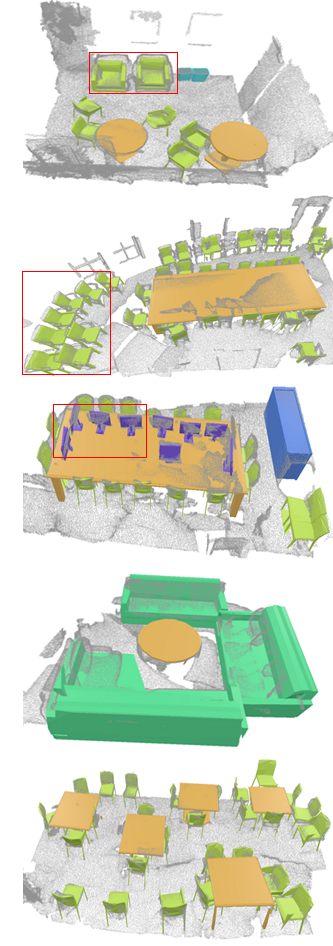}
        \label{Ours}
    }
    \quad
    \subfigure[Ground truth]{
        \includegraphics[width=0.24\textwidth]{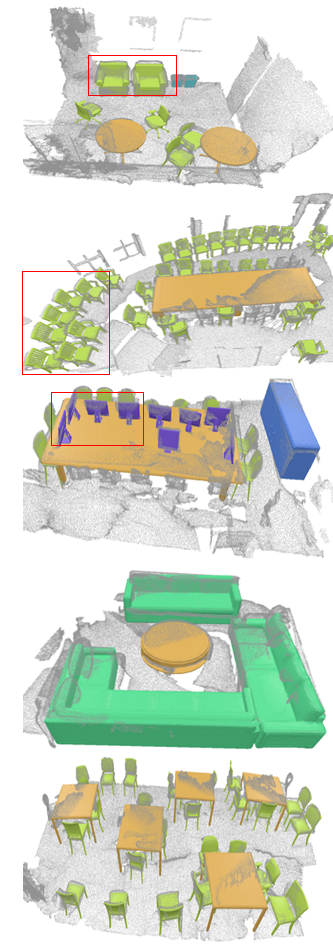}
        \label{Ground truth}
    }
    \caption{Qualitative comparison on the ScanNet dataset}
    \label{QC}
\end{figure*}

Table \ref{prec} reports the precision of object recognition at different IoU thresholds. The Prec@0.25 and Prec@0.5 are both improved more than 10 points by our DOCTR, indicating that our model greatly reduces false positive proposals. This shows that our method enables direct sparse predictions, and even without using NMS, it yields fewer false positives compared to existing methods with NMS.

\subsection{Ablation Study} 
We conduct ablation studies to analyze the effectiveness of our method on ScanNet. As shown in Table \ref{ablation}, the first row shows the performance of a baseline model that is modified from Mask3D \cite{Schult23ICRA} by directly adding box and shape prediction heads. Both the IoU@0.5 and CD@0.1 metrics of this baseline are inferior to those of DIMR, which shows that straight forward applying an object-centric Transformer-based model is not sufficient.
Adding hybrid bipartite matching module to the baseline model brings an improvement in the CD@0.1 metric, but the IoU@0.5 metric shows a slight decrease.
As shown in the third row, our proposed SGDQ for DTD layers achieves apparent improvement in both metrics compared to baseline model, and both metrics surpass those of DIMR. By combining the SGDQ design with hybrid bipartite matching, both metrics are further improved and reach the best results.


\begin{table}[htbp]
	\centering
	\fontsize{9}{10}\selectfont
	\begin{threeparttable}
\begin{tabular}{ccc|ccc}
			\toprule
			\multirow{1}{*}{}\makecell[c]{SGDQ} & \makecell[c]{Hybrid  Matching}& \makecell[c]{MEBR}
			&IoU@0.25&CD@0.1\cr
			\midrule
            \XSolidBrush&\XSolidBrush&\XSolidBrush&52.59&52.51\cr
			 \XSolidBrush&\Checkmark&\XSolidBrush&52.51&53.31\cr
			 \Checkmark&\XSolidBrush&\XSolidBrush&53.93&54.79\cr
			 \Checkmark&\Checkmark&\XSolidBrush&55.18&56.82\cr
			 \Checkmark&\Checkmark&\Checkmark&58.25&59.61\cr
			\bottomrule
		\end{tabular}
	\end{threeparttable}
 \caption{Ablation Study.  \label{ablation}}
\end{table}
\subsection{Qualitative Comparisons}
We present a qualitative comparison in Figure \ref{QC}. The version of our model to yield the results is the DOCTR (with MEBR). The first two rows in Figure \ref{QC} demonstrate that DIMR tends to yield erroneous segmentation when there are multiple objects in close proximity, resulting in multiple repetitive false positives. In comparison, our method achieves accurate segmentation and correct object reconstruction even for the closely placed objects. 
The third row shows that DIMR predicts incorrect orientations for the monitors and the fifth row shows that DIMR predicts some noticeable angular errors of the tables. In contrast, our method predicts correct results. 
Our method also outperforms DIMR in object shapes, such as for distinguishing between square and round tables in the fifth row. Moreover, DIMR is less accurate in predicting objects of limited presence in training set such as L-shaped sofas in the fourth row. 


As the visualization shows, our DOCTR demonstrates significant and consistent improvements than DIMR in the reconstructed scene in terms of object instance segmentation, pose accuracy, and shape quality.  


\section{Conclusion}
This paper introduces a novel Disentangled Object-Centric TRansformer (DOCTR) that allows learning with multiple objects for multiple sub-tasks in a unified manner. 
In particular, our design of semantic-geometry disentangled query is proved to be effective to improve performance on the different semantic and geometric sub-tasks. Extensive experiments demonstrate our superior performance than previous SOTA methods. At present, our method relies on a pretrained shape decoder to generate object meshes, which can be integrated within the main network for collaborative optimization in future work.
We hope our work inspires more future works to explore unified learning models to understand rich attributes of multiple objects in complex scenes.  


\section{Acknowledgments}
We would like to acknowledge contributions through helpful discussions from Hui Zhang and Yi Zhou.
\bibliography{aaai24}

\end{document}